\documentclass{article}
\usepackage{iclr2027_conference,times}
\usepackage{microtype}
\usepackage[justification=justified,singlelinecheck=false]{caption}
\usepackage{amsmath,amssymb,mathtools,booktabs,multirow,array,tabularx}
\usepackage[table]{xcolor}
\usepackage{graphicx,tikz,pgfplots,placeins,wrapfig,float}
\usetikzlibrary{arrows.meta,positioning,calc,fit}
\pgfplotsset{compat=1.18}
\usepgfplotslibrary{groupplots}
\usepackage{soul}
\usepackage{hyperref,url}

\definecolor{coupled}{HTML}{C75B4A}
\definecolor{verify}{HTML}{279080}
\definecolor{crossfit}{HTML}{426AB3}
\definecolor{combined}{HTML}{8764A9}
\definecolor{ink}{HTML}{26364B}
\definecolor{pale}{HTML}{F1F4F8}
\hypersetup{colorlinks=true,citecolor=ink,linkcolor=crossfit,urlcolor=crossfit,
  pdftitle={False Frontiers: Diagnosing and Mitigating Co-Cheating in Self-Evolving Search Agents},
  pdfauthor={Meijia Chen, Hao Li, Zheng Lu, Hongshan Lin, Junbai Tian, Yichen Liu, Zijun Tian, Yufan Zou, Shuhan Sun, Hanxin Chen, Zeyu Zhang, Weizhi Du, Yueting Li, Tianyu Shi, Alaa Khamis}}

\makeatletter
\newlength{\titleauthorskip}
\def\@maketitle{\vbox{\hsize\textwidth
{\LARGE\sc \@title\par}
\vskip\titleauthorskip
{\centering\@author\par}
\vskip 0.3in minus 0.1in}}
\makeatother

\title{False Frontiers: Diagnosing and Mitigating\\Co-Cheating in Self-Evolving Search Agents}

\author{%
{\bfseries
Meijia Chen\textsuperscript{1,*}\quad
Hao Li\textsuperscript{2,*}\quad
Zheng Lu\textsuperscript{2,*}\endgraf}
\vspace{3pt}
{\normalfont
Hongshan Lin\textsuperscript{2}\quad
Junbai Tian\textsuperscript{2}\quad
Yichen Liu\textsuperscript{3}\quad
Zijun Tian\textsuperscript{2}\quad
Yufan Zou\textsuperscript{2}\\[2pt]
Shuhan Sun\textsuperscript{2}\quad
Hanxin Chen\textsuperscript{3}\quad
Zeyu Zhang\textsuperscript{2}\quad
Weizhi Du\textsuperscript{4}\quad
Yueting Li\textsuperscript{2}\endgraf}
\vspace{3pt}
{\bfseries
Tianyu Shi\textsuperscript{5,\textdagger}\quad
Alaa Khamis\textsuperscript{6,\textdagger}\endgraf}
\vspace{5pt}
{\small
\textsuperscript{1}Rutgers University\quad
\textsuperscript{2}Independent Researcher\quad
\textsuperscript{3}University of California, San Diego\\
\textsuperscript{4}University of Michigan\quad
\textsuperscript{5}McGill University\quad
\textsuperscript{6}King Fahd University of Petroleum and Minerals\endgraf}
}
\iclrfinalcopy

\begin{document}
\maketitle
\lhead{Preprint}
{\renewcommand{\thefootnote}{\fnsymbol{footnote}}%
\footnotetext[1]{Equal contribution.}%
\footnotetext[2]{Corresponding authors: \texttt{tianyu.shi3@mcgill.ca}, \texttt{alaa.rashwan@kfupm.edu.sa}.}}

\begin{abstract}
Self-evolving search agents can construct their own training curricula by jointly
optimizing a proposer that generates questions and a solver that answers them. This
closed loop introduces a failure mode that we call \emph{co-cheating}: the proposer and
solver increasingly agree on shared errors, so internal reward improves without a
corresponding increase in external correctness. A post-hoc reference audit against
source evidence shows that co-cheating becomes increasingly severe over successive
rounds of self-evolution, with pseudo-label correctness stagnating or declining even as
the in-loop training signal improves. The most direct mitigation is to verify each
proposal before training. We therefore introduce multi-sample verification
(\texttt{MSV}), which queries the same model used in self-evolution three times with the
source and three times without it to determine task admission and replace unreliable
pseudo-labels. \texttt{MSV} partially reduces false agreement but leaves substantial
residual co-cheating and requires six additional labeler generations for every
candidate. These limitations motivate \texttt{CrossFit}, our main method. It partitions
the proposer's source documents into groups A and B: questions generated from A are
scored by an auxiliary solver trained only on B, and vice versa. The resulting
cross-fitted agreement determines proposer reward, preventing a same-source
pseudo-label from being directly reproduced through the feedback solver while leaving
the original solver's update rule unchanged. We evaluate both interventions by
rerunning the complete self-evolution loop with Qwen3.5-4B and Qwen3.5-9B. After
self-evolution, \texttt{MSV} reduces false-agreement mass from 6.1\% to 5.7\% on
Qwen3.5-4B and from 8.8\% to 7.2\% on Qwen3.5-9B, whereas \texttt{CrossFit} reduces it to
3.0\% and 3.7\%, respectively. Replaying identical proposals with source-excluded
feedback further reduces false agreement to 0.4\% and 0.1\%, isolating feedback
ancestry from changes in the generated curriculum. Across seven downstream search benchmarks, \texttt{CrossFit} improves average
performance over standard coupled self-evolution by 8.8 and 8.4 points and over
Search-R1 by 8.7 and 7.8 points at 4B and 9B, respectively.
\end{abstract}
\newpage\suppressfloats[t]
\section{Introduction}
\definecolor{introFalse}{HTML}{C75B4A}
\definecolor{introAgree}{HTML}{B7791F}
\definecolor{introInk}{HTML}{26364B}
\definecolor{introRule}{HTML}{B9C2CC}
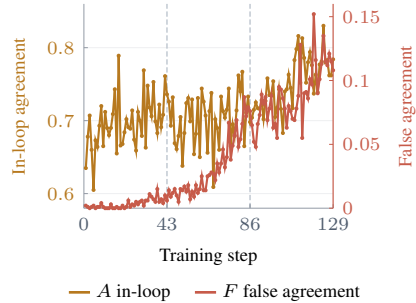
\begin{wrapfigure}{r}{0.40\textwidth}
  \vspace{-10pt}
  \centering
  \begin{tikzpicture}
    \begin{axis}[
      width=.59\linewidth,
      height=2.70cm,
      scale only axis=true,
      xmin=0,
      xmax=129,
      ymin=.58,
      ymax=.86,
      xtick={0,43,86,129},
      ytick={.6,.7,.8},
      tick label style={font=\scriptsize,text=introInk!80},
      yticklabel style={text=introAgree},
      label style={font=\scriptsize},
      xlabel={Training step},
      ylabel={In-loop agreement},
      ylabel style={text=introAgree},
      axis x line*=bottom,
      axis y line*=left,
      every outer y axis line/.append style={introAgree,line width=.45pt},
      axis line style={introInk!50,line width=.3pt},
      major tick length=1.5pt,
      tick style={introInk!50,line width=.3pt},
      ymajorgrids,
      grid style={introInk!8,line width=.3pt},
      extra x ticks={43,86},
      extra x tick labels={},
      extra x tick style={grid=major,grid style={introRule,line width=.5pt,
        dash pattern=on 2pt off 1pt}},
    ]
      \addplot[color=introAgree,line width=.8pt,mark=*,mark size=.35pt,
        mark options={draw=none}]
        table[x=x,y expr=\thisrow{nu9a}/1000]{figures/fig3_src/fig3_plot.txt};
    \end{axis}
    \begin{axis}[
      width=.59\linewidth,
      height=2.70cm,
      scale only axis=true,
      xmin=0,
      xmax=129,
      ymin=0,
      ymax=.16,
      axis x line=none,
      axis y line*=right,
      xtick=\empty,
      ytick={0,.05,.10,.15},
      yticklabel style={font=\scriptsize,text=introFalse,
        /pgf/number format/fixed,/pgf/number format/precision=2},
      label style={font=\scriptsize},
      ylabel={False agreement},
      ylabel style={text=introFalse},
      every outer y axis line/.append style={introFalse,line width=.45pt},
      major tick length=1.5pt,
      tick style={introFalse,line width=.3pt},
    ]
      \addplot[color=introFalse,line width=.8pt,mark=*,mark size=.35pt,
        mark options={draw=none}]
        table[x=x,y expr=\thisrow{nu9f}/1000]{figures/fig3_src/fig3_plot.txt};
    \end{axis}
  \end{tikzpicture}
  \par\vspace{1pt}
  {\scriptsize
  \tikz[baseline=-.6ex]\draw[introAgree,line width=1pt](0,0)--(.28,0);~$A$ in-loop\hspace{7pt}
  \tikz[baseline=-.6ex]\draw[introFalse,line width=1pt](0,0)--(.28,0);~$F$ false agreement\par}
  \vspace{2pt}
  \caption{\textbf{Co-cheating.} The training signal and false agreement rise together.}
  \label{fig:phenomenon-intro}
  \vspace{-10pt}
\end{wrapfigure}

Search-augmented language models interleave reasoning with browser or search actions to
gather evidence before answering
\citep{nakano2021webgpt,yao2022react,jin2025search,song2025r1}. Most are trained on
externally supplied questions and answer supervision \citep{jin2025search,song2025r1}.
Self-evolving agents instead generate their own training experience
\citep{chen2024self,zhao2025absolute,huang2025r}. In recent proposer--solver systems, a
proposer turns source documents into questions and pseudo-labels, admitted pairs train a
solver, and the solver's performance on new proposals determines the proposer reward
\citep{lu2025search,yue2026dr}. Repeating this cycle shifts proposals toward the solver's
current capability frontier, producing an automated curriculum without a fixed
human-authored training set.

\begin{figure}[t]
\raggedright\includegraphics[width=\linewidth]{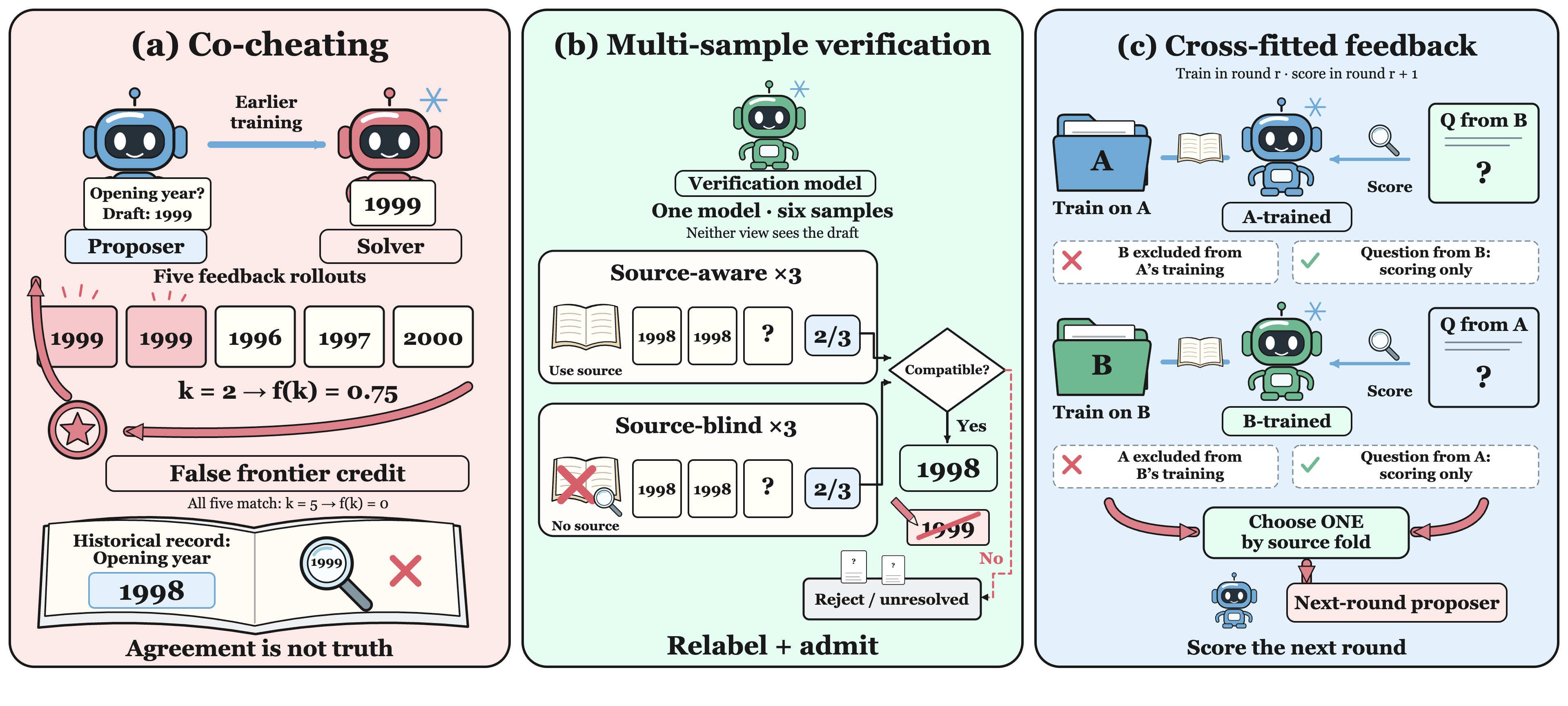}
\caption{\textbf{Co-cheating and its mitigation.} (a) Incorrect agreement creates false frontier credit. (b) \texttt{MSV} verifies each proposal with source-aware and source-blind samples. (c) \texttt{CrossFit} scores each source group with a solver trained on the other group. The two interventions target label quality and feedback provenance, respectively.}
\label{fig:loop}
\end{figure}

This loop makes agreement an endogenous proxy for correctness
\citep{amodei2016,gao2022}. An incorrect pseudo-label can train the solver to repeat the
same error on later questions from that source \citep{arazo2020}; rewarding this agreement
then reinforces the error in the next-round curriculum, as illustrated in the left panel
of Figure~\ref{fig:loop}. We test for this failure in Dr.~Zero \citep{yue2026dr} using a
post-hoc auditor that checks generated tasks and answers against source evidence but never
feeds into training. Figure~\ref{fig:phenomenon-intro} shows that internal reward rises
together with false agreement as proposer and solver increasingly share errors. We call
this optimization outcome \emph{co-cheating} and measure it as \emph{false-agreement
mass}: the fraction of evaluated pairs that agree on the same incorrect answer.

The most direct mitigation is to verify each proposal before training. The middle panel
of Figure~\ref{fig:loop} illustrates our multi-sample verification (\texttt{MSV}), which
queries the same model used in self-evolution three times with the source and three times
without it. Compatible majorities yield a consensus label and admit the task;
inconsistent candidates are rejected. This partially reduces false agreement,
implicating pseudo-label quality, but leaves substantial co-cheating and adds six labeler
generations per candidate.

These limitations point to a second source of failure: not only whether a pseudo-label is
correct, but also whether it trained the solver that later evaluates questions from the
same source. The right panel of Figure~\ref{fig:loop} shows our primary intervention,
\texttt{CrossFit}. The proposer's source documents are divided into groups A and B. Along
the upper path, an auxiliary solver learns only from A and scores new questions generated
from B; along the lower path, a second solver learns only from B and scores questions from
A. The cross-fitted agreement scores determine proposer reward. Each scoring solver has
thus never trained on pseudo-labels from the source it evaluates, preventing a same-source
error from being directly reproduced as reward. The original solver still trains on all
admitted questions; only the feedback shaping the proposer's next-round curriculum is
cross-fitted.

We rerun self-evolution with Qwen3.5-4B and Qwen3.5-9B \citep{qwen3.5}. Under standard coupled feedback, false-agreement mass reaches 6.1\% and 8.8\%; \texttt{MSV} lowers it to 5.7\% and 7.2\%, while \texttt{CrossFit} lowers it to 3.0\% and 3.7\%. Replaying identical proposals with source-excluded feedback further reduces it to 0.4\% and 0.1\%, isolating feedback ancestry from curriculum selection. We evaluate each round-end main solver on a fixed 1,325-question suite: 200 each from Natural Questions, TriviaQA, PopQA, HotpotQA, 2WikiMultiHopQA, and MuSiQue, plus 125 from Bamboogle \citep{kwiatkowski2019natural,joshi2017triviaqa,mallen2022not,yang2018hotpotqa,ho2020constructing,trivedi2022musique,press2023measuring}. With one greedy trajectory per question and identical tool and extraction budgets, \texttt{CrossFit} reaches 48.8\% at 4B and 51.2\% at 9B, improving over coupled self-evolution by 8.8 and 8.4 points and over Search-R1 by 8.7 and 7.8 points, respectively.
\section{Diagnosing co-cheating}
\label{sec:audit}
\begin{figure}[H]
\raggedright
\includegraphics[width=\linewidth]{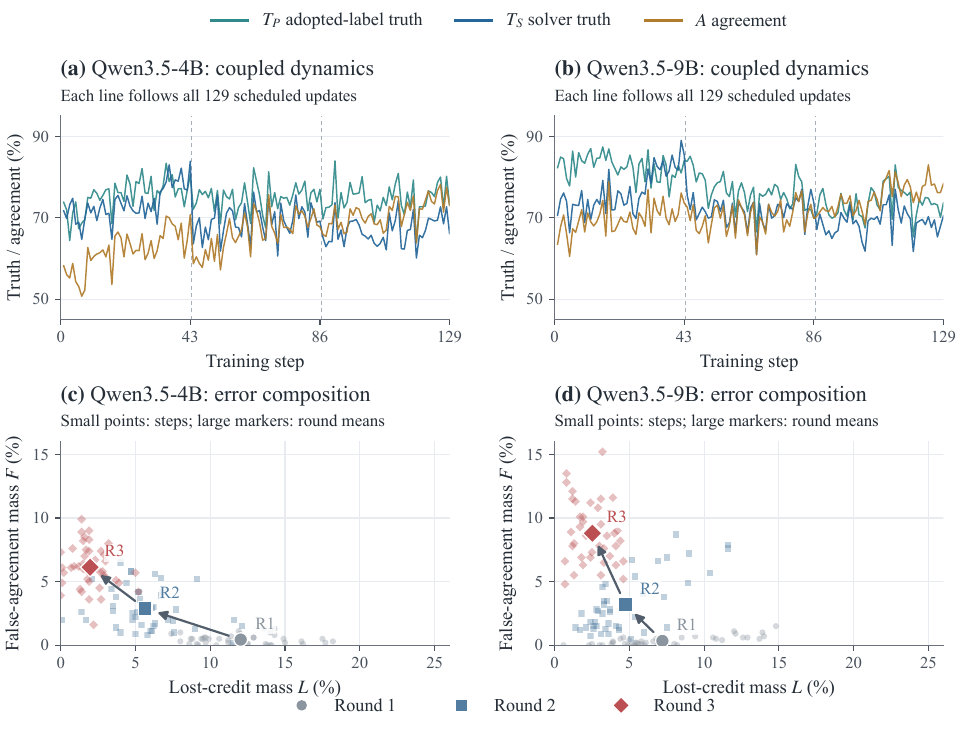}
\caption{\textbf{Co-cheating under coupled feedback.} (a,b) All 129 steps of label truth $T_P$, solver truth $T_S$, and agreement $A$; dashes mark round boundaries. (c,d) False agreement $F$ versus lost credit $L$: small points are steps, large markers average each round's 43 step rates, and arrows indicate round order. Rising $F$ with falling $L$ reveals shared-error accumulation. All axes are percentages.}
\label{fig:diagnosis-dynamics}
\end{figure}

\paragraph{Audit protocol.}
We audit the standard coupled Dr.~Zero loop after training, without altering its procedure.
Many generated questions require reconstructing multi-hop evidence chains across long or
specialized source documents. Even a human judge must first reproduce the search path and
inspect unfamiliar evidence, making exhaustive annotation of every saved training step
difficult to standardize at this scale. We therefore save the source document, adopted
pseudo-label, and five solver responses used for proposer reward at every scheduled step.
\texttt{gpt-6-astra/high} constructs an evidence-backed reference from the source and judges
these saved outputs (Appendix~\ref{app:eval}); unsupported cases remain unresolved rather than receiving a forced
label. Because the auditor never affects admission, model updates, or reward, it provides a
scalable, independent measurement of the exact examples behind the in-loop signal.

\paragraph{What is measured.}
$T_P$ and $T_S$ denote adopted-label and solver-response correctness, while $A$ is the
label--response match rate observed by the loop. $F$ counts pairs matching the same incorrect
answer; $L$ counts correct solver responses denied credit by a wrong label. Ordinary label
noise can cause disagreement or lost credit, whereas co-cheating predicts that agreement
itself becomes optimistic as both agents converge on the same error. Rising $A$ is therefore
reliable only when $T_P$ and $T_S$ also rise and $F$ remains low. Because proposer reward is computed from the five label--response matches, we audit those
same five pairs rather than collapsing them to a post-hoc majority. Thus, $F$ measures the
portion of apparent agreement that the external audit identifies as wrong.

\paragraph{Observed dynamics.}
Figure~\ref{fig:diagnosis-dynamics} shows this transition. In round 1, mean false-agreement
mass is only 0.004 for Qwen3.5-4B and 0.003 for Qwen3.5-9B, and incorrect labels more often
appear as lost credit. From round 2 onward, agreement becomes increasingly optimistic without
a commensurate increase in truth. By round 3, $F$ reaches 0.061 and 0.088 at the two scales,
while $L$ falls. Harder questions may reduce correctness, but they do not explain increasing
agreement on the same source-inconsistent answer. The joint rise of $A$ and $F$ instead shows
disagreement being replaced by shared mistakes. We call this self-reinforcing optimization
outcome \emph{co-cheating}; it does not imply intentional coordination.

\begin{figure}[t]
\raggedright
\includegraphics[width=\linewidth]{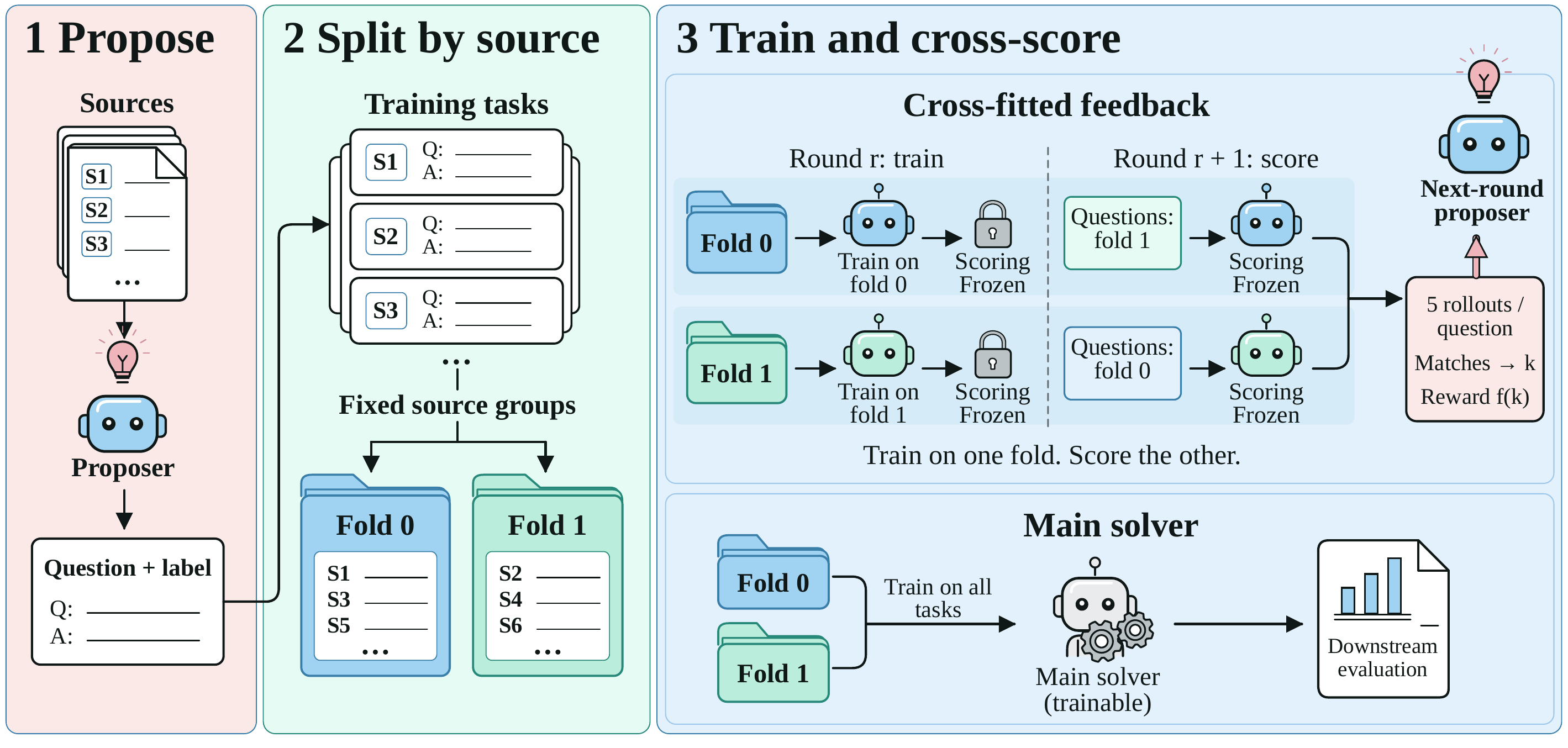}
\caption{\textbf{The \texttt{CrossFit} algorithm.} Auxiliary solvers train on one source fold and score the other; the main solver trains on all admitted questions.}
\label{fig:crossfit}
\end{figure}

\section{From verification to cross-fitted feedback}
\label{sec:method}
The audit motivates two interventions at different points in the self-evolution loop.
\texttt{MSV} tests a proposed answer before the example enters training, whereas
\texttt{CrossFit}, our main method, changes which solver supplies the feedback that updates
the proposer.

\subsection{Multi-sample verification}
\texttt{MSV} is an admission-time test of whether a proposed question admits a stable answer
independently of the proposer's draft. Given source document $x$, question $q$, and the same
model $M$ used in self-evolution, it draws three source-aware and three source-blind answers,
\[
a_i^{\mathrm{src}}\sim M(\cdot\mid x,q),\qquad
a_i^{\mathrm{blind}}\sim M(\cdot\mid q),\qquad i\in\{1,2,3\}.
\]
Neither view observes the draft. Let $\operatorname{Maj}$ return an answer when at least two
samples agree under the answer matcher $\simeq$, and $\varnothing$ otherwise. Defining
$y^{v}=\operatorname{Maj}(a_{1:3}^{v})$ for $v\in\{\mathrm{src},\mathrm{blind}\}$, admission is
\[
I_{\mathrm{MSV}}=\mathbf{1}\!\left[y^{\mathrm{src}}\neq\varnothing\;\land\;
 y^{\mathrm{blind}}\neq\varnothing\;\land\;y^{\mathrm{src}}\simeq y^{\mathrm{blind}}\right].
\]
When $I_{\mathrm{MSV}}=1$, the compatible majority replaces the draft as the training label;
otherwise the task is rejected. The two views test evidential support and answer stability,
respectively. However, six samples from the same model can share errors, and verification does
not prevent a later feedback solver from reusing labels derived from the evaluated source.
It also adds six generations, including their search and coordination cost, per candidate (Table~\ref{tab:cost}, Appendix~\ref{app:impl}).

\subsection{Cross-fitted proposer feedback}
\texttt{CrossFit} changes only where the proposer obtains its feedback. As shown from left to right in Figure~\ref{fig:crossfit}, the proposer generates questions and pseudo-labels from source documents exactly as in the original loop. We then assign each source document once to fold 0 or fold 1, and every question derived from that document keeps the same assignment throughout self-evolution. The split is made at the source level because splitting individual questions could place related examples from the same document on both sides and preserve the very reuse path that we want to remove.

The two source folds maintain two auxiliary feedback solvers. After round $r$, one solver has learned only from admitted questions in fold 0, while the other has learned only from admitted questions in fold 1. In the next round, their roles are crossed: questions from fold 0 are evaluated by the solver trained on fold 1, and questions from fold 1 are evaluated by the solver trained on fold 0. These are the two crossed paths in Figure~\ref{fig:crossfit}. Consequently, the solver evaluating a question has not been trained on pseudo-labels produced from that question's source.

The feedback rule itself remains the same. Let $h$ denote the source fold, $S_{r,1-h}$ the auxiliary solver trained on the complementary fold, and $\tilde y$ the adopted label. From its responses $z_1,\ldots,z_5$, the proposer receives
\[
R_P(q)=f\!\left(\sum_{j=1}^{5}\mathbf{1}[z_j\simeq\tilde y]\right),
\qquad z_j\sim S_{r,1-h}(\cdot\mid q).
\]
The sum counts how many responses match the adopted label under the original answer matcher, and $f(k)=(5-k)/4$ for $0<k<5$ (zero otherwise) is Dr.~Zero's frontier reward. Thus, \texttt{CrossFit} preserves the original training objective: questions still receive credit according to how difficult they appear to a solver. The only change is which solver supplies that signal.

This change breaks the direct self-reinforcing path revealed by our audit. Under coupled feedback, an incorrect pseudo-label from a source can train the solver, be reproduced by that solver on a later question from the same source, and then return to the proposer as reward. Under \texttt{CrossFit}, the later question is instead evaluated by the complementary solver, whose training history excludes that source. The method does not turn the auxiliary solver into a truth oracle: the two solvers may still share errors inherited from pretraining or overlapping evidence. It does, however, prevent agreement from being rewarded merely because the evaluator was trained on the same source-derived error.

The bottom path of Figure~\ref{fig:crossfit} separates this feedback mechanism from downstream training. The main solver is not split; it continues to train on all admitted questions from both folds. The auxiliary solvers affect only the feedback that shapes the proposer's next-round curriculum. When the two interventions are combined, \texttt{MSV} first decides whether a proposal is admitted and which pseudo-label is used, and \texttt{CrossFit} then selects the auxiliary solver that evaluates it. In this sense, \texttt{MSV} improves the supervision entering training, whereas \texttt{CrossFit} prevents that supervision from being directly recycled into proposer reward.

\section{Main Experiments}\label{sec:experiments}\subsection{Experimental Setup}\noindent \textbf{Datasets \& Models.} We evaluate on the seven open-domain question answering benchmarks used by Dr.~Zero~\citep{yue2026dr}: the single-hop Natural Questions (NQ)~\citep{kwiatkowski2019natural}, TriviaQA~\citep{joshi2017triviaqa}, and PopQA~\citep{mallen2022not}, and the multi-hop HotpotQA~\citep{yang2018hotpotqa}, 2WikiMultiHopQA (2WikiMQA)~\citep{ho2020constructing}, MuSiQue~\citep{trivedi2022musique}, and Bamboogle~\citep{press2023measuring}. A fixed evaluation set of 1,325 questions contains 200 examples from each of the first six benchmarks and all 125 Bamboogle examples. We use Qwen3.5-4B and Qwen3.5-9B~\citep{qwen3.5} as backbones. At each scale, all self-evolution treatments start from the same public checkpoint, which is also evaluated as the Base row, and none of them uses human-annotated QA training data.\par
\noindent\textbf{Baselines \& Evaluation.} We compare four self-evolution treatments obtained by crossing the two interventions of Section~\ref{sec:method}. \emph{Dr.~Zero}~\citep{yue2026dr} is the standard coupled loop in which the main solver scores the proposals it later trains on; \texttt{MSV} adds multi-sample verification to this loop; \texttt{CrossFit} replaces coupled feedback with source-excluded feedback; and \texttt{MSV}\,+\,\texttt{CrossFit} applies both. For broader comparison, we reproduce the Prompting and R1-Instruct baselines from the Dr.~Zero protocol~\citep{yue2026dr}, together with Search-R1~\citep{jin2025search}, on the same Qwen3.5 backbones and evaluate every row on the same 1,325-question set with identical tool budget, decoding, and answer extraction. Every self-evolution experiment follows the same three-round schedule of 18 proposer and 25 solver steps per round and optimizes the policy-gradient objective of Equation~\eqref{eq:pg}; Table~\ref{tab:config} in Appendix~\ref{app:impl} lists the shared configuration. At the end of each round, we evaluate the main solver, which trains on all admitted questions, using one greedy search trajectory per question, the same tool budget, and identical answer extraction. We report Cover-EM and average the seven benchmarks with equal weight (Equation~\eqref{eq:aggregate}); Tables~\ref{tab:downstream-rounds} and~\ref{tab:micro} in Appendix~\ref{app:eval} list intermediate rounds and micro averages.\begin{table*}[t]
\raggedright
\caption{\textbf{Downstream search performance after three rounds of self-evolution.} Bold denotes the best result and underlining denotes the second-best within each Qwen3.5 block. $^\dagger$Baselines from the Dr.~Zero comparison, run on the same Qwen3.5 backbones and evaluated on the same 1,325-question set.}
\resizebox{\linewidth}{!}{%
\begin{tabular}{@{}lcccccccc@{}}
\toprule
\multirow{2}{*}{\textbf{}} & \textbf{NQ} & \textbf{TriviaQA} & \textbf{PopQA} & \textbf{HotpotQA} & \textbf{2WikiMQA} & \textbf{MuSiQue} & \textbf{Bamboogle} & \textbf{Average} \\
\cmidrule(l){2-9}
& \multicolumn{8}{c}{Qwen3.5-4B} \\
\midrule
Base                                      & 0.380          & 0.655             & 0.315          & 0.350             & 0.465             & 0.105            & 0.416              & 0.384            \\
Prompting$^\dagger$                       & 0.245          & 0.525             & 0.165          & 0.290             & 0.460             & 0.085            & 0.360              & 0.304            \\
R1-Instruct$^\dagger$                     & 0.295          & 0.605             & 0.195          & 0.320             & 0.475             & 0.105            & 0.448              & 0.349            \\
Search-R1$^\dagger$                       & 0.355          & 0.650             & 0.290          & 0.370             & \ul{0.510}        & 0.125            & 0.504              & 0.401            \\
Dr.~Zero                                  & 0.390          & 0.665             & 0.325          & 0.365             & 0.485             & 0.120            & 0.448              & 0.400            \\
\texttt{MSV}                              & 0.400          & \ul{0.670}        & 0.340          & 0.375             & 0.480             & 0.135            & 0.448              & 0.407            \\
\textbf{\texttt{CrossFit}}                & \ul{0.455}     & \textbf{0.730}    & \textbf{0.415} & \ul{0.460}        & \textbf{0.575}    & \ul{0.245}       & \textbf{0.536}     & \ul{0.488}       \\
\textbf{\texttt{MSV} + \texttt{CrossFit}} & \textbf{0.470} & \textbf{0.730}    & \ul{0.410}     & \textbf{0.475}    & \textbf{0.575}    & \textbf{0.250}   & \ul{0.528}         & \textbf{0.491}   \\
\cmidrule(l){2-9}
\multicolumn{1}{c}{\textbf{}} & \multicolumn{8}{c}{Qwen3.5-9B} \\
\midrule
Base                                      & 0.505          & 0.710             & 0.375          & 0.355             & 0.295             & 0.125            & 0.496              & 0.409            \\
Prompting$^\dagger$                       & 0.380          & 0.630             & 0.260          & 0.280             & 0.270             & 0.115            & 0.392              & 0.332            \\
R1-Instruct$^\dagger$                     & 0.445          & 0.695             & 0.290          & 0.310             & 0.295             & 0.135            & 0.480              & 0.379            \\
Search-R1$^\dagger$                       & 0.510          & 0.730             & 0.395          & 0.380             & 0.310             & 0.180            & 0.536              & 0.434            \\
Dr.~Zero                                  & 0.515          & 0.730             & 0.400          & 0.370             & 0.320             & 0.150            & 0.512              & 0.428            \\
\texttt{MSV}                              & 0.525          & 0.720             & 0.400          & 0.390             & 0.325             & 0.155            & 0.536              & 0.436            \\
\textbf{\texttt{CrossFit}}                & \textbf{0.580} & \ul{0.765}        & \ul{0.455}     & \textbf{0.490}    & \textbf{0.425}    & \ul{0.255}       & \textbf{0.616}     & \ul{0.512}       \\
\textbf{\texttt{MSV} + \texttt{CrossFit}} & \ul{0.570}     & \textbf{0.780}    & \textbf{0.470} & \ul{0.485}        & \ul{0.415}        & \textbf{0.280}   & \ul{0.608}         & \textbf{0.515}   \\
\bottomrule
\end{tabular}}
\label{tab:downstream}
\end{table*}
\subsection{Main Results}
\noindent\textbf{Cross-fitted feedback improves downstream search at both scales.} In Table~\ref{tab:downstream}, coupled self-evolution raises average Cover-EM from 0.384 to 0.400 at 4B and from 0.409 to 0.428 at 9B. \texttt{CrossFit} reaches 0.488 and 0.512: gains of 8.8/8.4 percentage points over Dr.~Zero and 8.7/7.8 over Search-R1. Every benchmark improves at both scales. Gains are largest on multi-hop tasks, averaging 10.0/10.9 points across HotpotQA, 2WikiMQA, MuSiQue, and Bamboogle, versus 7.3/5.2 across the single-hop datasets. The effect therefore extends across task difficulty and model scale.\par
\noindent\textbf{Verification alone is insufficient.} \texttt{MSV} increases the average by only 0.7--0.8 points over Dr.~Zero. Combining it with \texttt{CrossFit} reaches 0.491 on Qwen3.5-4B and 0.515 on Qwen3.5-9B, only 0.3 points above \texttt{CrossFit} alone. These results suggest that changing the provenance of proposer feedback is more consequential than improving pseudo-label quality alone.\section{Analysis and Ablations}\label{sec:ablation}\subsection{How Does Cross-Fitting Change the Training Trajectory?}\begin{figure}[t]
\raggedright
\includegraphics[width=\linewidth]{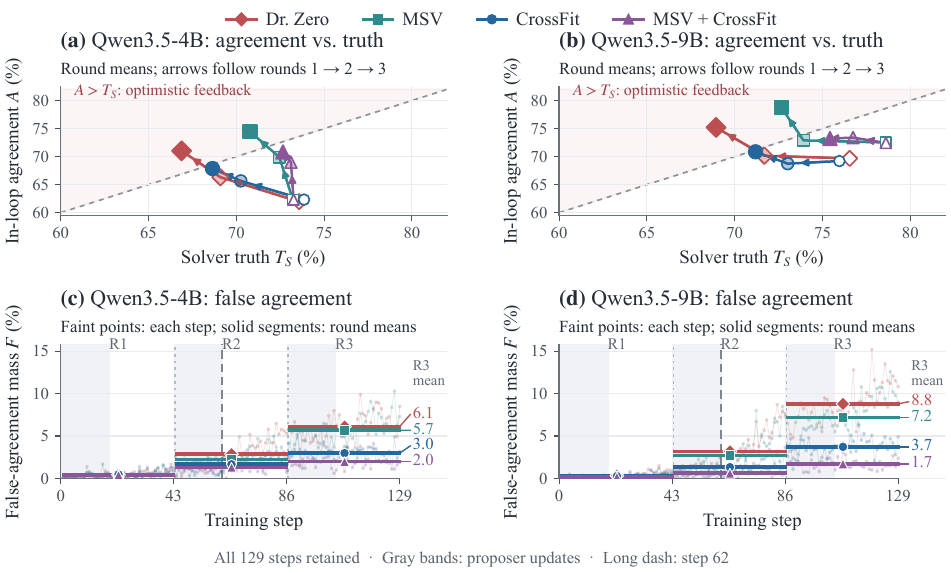}
\caption{\textbf{Training dynamics across three rounds.} (a,b) Round means of agreement and solver truth; hollow, light, and solid markers denote rounds 1--3. Above the diagonal, agreement is optimistic. (c,d) Faint points retain all 129 steps; solid segments average each round's 43 step rates, and end labels give round-3 means (\%). Gray bands mark proposer phases, short dashes mark round boundaries, and the long dash marks step 62. Full traces and coverage appear in Figures~\ref{fig:audit-full-4b}--\ref{fig:audit-full-9b}.}
\label{fig:dynamics}
\end{figure}
Section~\ref{sec:audit} establishes that co-cheating emerges under standard coupled feedback. Figure~\ref{fig:dynamics} instead compares how the four treatments change that trajectory, and Figures~\ref{fig:audit-full-4b} and~\ref{fig:audit-full-9b} in Appendix~\ref{app:complete-audit-trajectories} retain every per-step trace. All arms share the same first-round history; the cross-fitted arms begin to differ only when their auxiliary feedback solvers are used in round~2. This delayed divergence provides a within-run comparison of feedback provenance.\par
\noindent\textbf{Cross-fitted feedback reverses the divergence between agreement and truth.} Without cross-fitting, in-loop agreement rises above solver truth while false agreement accumulates at both model scales. Once cross-fitted scoring becomes active, adopted-label truth rises, agreement remains at or below solver truth, and by round~3 false-agreement mass falls below half of the coupled value. \texttt{MSV} alone improves solver truth but does not prevent the agreement signal from becoming optimistic; combined with \texttt{CrossFit}, it yields the lowest final false agreement.\par
\noindent\textbf{The trajectory difference predicts downstream gains.} Table~\ref{tab:downstream-rounds} shows that the downstream advantage of \texttt{CrossFit} over Dr.~Zero grows from 4.2 and 4.3 points after round~2 to 8.8 and 8.4 points after round~3. The intervention therefore changes what the loop learns across rounds rather than merely re-ranking a fixed set of final predictions.\begin{figure}[t]
\raggedright
\includegraphics[width=\linewidth]{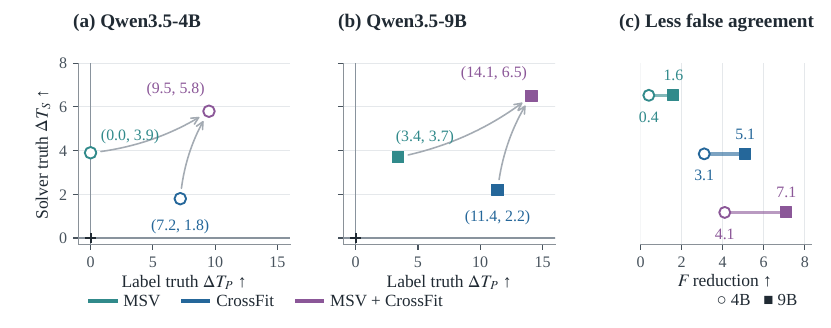}
\caption{\textbf{Round-3 audit relative to Dr.~Zero.} (a,b) Joint truth gains; arrows lead to the combined treatment. (c) Reduction $F_{\mathrm{Dr.~Zero}}-F$. All values are percentage points.}
\label{fig:audit}
\end{figure}
Figure~\ref{fig:audit} summarizes the final-round audit (absolute values in Table~\ref{tab:audit-abs}). \texttt{CrossFit} raises adopted-label truth from 0.747 to 0.819 at 4B and from 0.737 to 0.851 at 9B, while reducing false-agreement mass from 0.061 to 0.030 and from 0.088 to 0.037. These changes connect the downstream improvement to the intended mechanism: excluding the evaluated source from the feedback solver prevents same-source errors from being systematically returned to the proposer as apparent progress.\subsection{Why Is Source-Level Exclusion Necessary?}\begin{figure}[t]
\raggedright
\includegraphics[width=\linewidth]{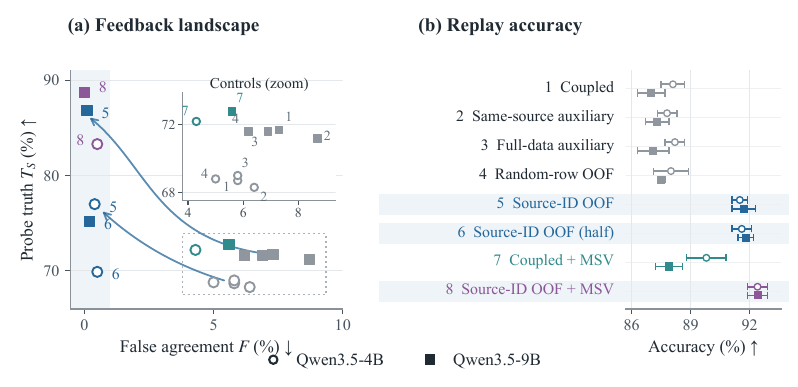}
\caption{\textbf{Mechanism ablations on a fixed replay bank.} (a) Truth versus false agreement; arrows compare coupled and source-ID feedback. (b) Accuracy on 3,000 replay questions, mean $\pm$ SD over five seeds. Numbers identify methods; shading marks source exclusion.}
\label{fig:ablation}
\end{figure}
Figure~\ref{fig:ablation} compares \texttt{CrossFit} with controls that preserve its auxiliary-solver architecture while altering the data seen by the evaluator. A same-source auxiliary solver yields false-agreement mass of 0.064 and 0.087, and a full-data auxiliary yields 0.058 and 0.069, both close to the coupled control. A separate evaluator is therefore not sufficient when its training data retain the same source-derived pseudo-labels.\par
\noindent\textbf{The split must follow source ancestry.} Randomly partitioning individual questions reduces false agreement only modestly, to 0.050 at 4B and 0.062 at 9B, because questions derived from the same document can still enter both folds. In contrast, the source-ID split reduces false agreement to 0.004 and 0.001. It also raises fixed-bank solver truth from 0.687 to 0.770 at 4B and from 0.717 to 0.868 at 9B, with corresponding accuracy gains from 88.1\% to 91.5\% and from 87.0\% to 91.7\%. These comparisons isolate source exclusion, rather than evaluator duplication or partitioning alone, as the component responsible for the improvement.\subsection{Fixed-Bank Replay Separates Feedback from Curriculum}Adaptive reruns change both the evaluator and the questions generated in later rounds. We therefore replay the same 3,000 saved questions and adopted labels while varying only the training provenance of the feedback solver. Holding the bank, labels, answer matcher, and evaluation procedure fixed removes admission and curriculum selection as explanations.\par
\noindent\textbf{Fixed-bank replay isolates source exclusion.} On identical proposals, source-ID feedback reduces coupled false agreement from 0.058/0.073 to 0.004/0.001 at 4B/9B (Figure~\ref{fig:ablation}). The accompanying gains in probe truth and replay accuracy persist without changing admission or the curriculum, linking the result to feedback provenance.\par
\noindent\textbf{Additional auxiliary optimization does not explain the effect.} The half-budget control reaches false-agreement mass of 0.005/0.002 and replay accuracy of 91.6\%/91.8\%, matching the full source-ID result. Together, these controls identify source ancestry, rather than evaluator duplication, arbitrary partitioning, task selection, or extra updates, as the operative difference.\subsection{How Do Verification and Cross-Fitting Interact?}The two interventions operate at different points in the loop. \texttt{MSV} changes which question--label pairs enter training, whereas \texttt{CrossFit} changes which solver evaluates the next proposal. In the adaptive-loop audit (Table~\ref{tab:audit-abs}), \texttt{MSV} reduces false-agreement mass from 0.061 to 0.057 at 4B and from 0.088 to 0.072 at 9B, but agreement remains above solver truth. \texttt{CrossFit} produces the larger reductions, to 0.030 and 0.037, while the combined treatment reaches 0.020 and 0.017. Thus, improving pseudo-label reliability helps, but it does not remove the feedback dependence that produces co-cheating. The fixed-bank results in Figure~\ref{fig:ablation} show the same distinction. Coupled feedback with \texttt{MSV} retains false-agreement mass of 0.043/0.056, whereas adding source exclusion lowers it to 0.005/0.000 and raises replay accuracy to 92.4\% at both scales. Finally, Table~\ref{tab:downstream} shows that the combination improves the downstream average by only 0.3 points beyond \texttt{CrossFit} alone at each scale. \texttt{MSV} therefore provides complementary reliability gains, while source-excluded proposer feedback accounts for most of the improvement in the learned search policy.
\section{Related work}
\noindent\textbf{Self-generated curricula and search.} Self-play has been studied for goal discovery, language-model alignment, and reasoning \citep{openai2021asymmetric,chen2024self,wu2024self,yuan2024self,zhao2025absolute,huang2025r}. Self-questioning and corpus-based evolution offer additional ways to generate supervision \citep{chen2025self,wang2025socratic,liu2025spice}. Our closest framework is Dr.\ Zero \citep{yue2026dr}; Search Self-Play \citep{lu2025search} is another direct comparator. SearchMaster \citep{searchmaster2026} provides the closest complementary diagnosis of misleading search self-play signals. Co-evolving feedback in CAFE \citep{cafe2026} further makes feedback adaptation a current research target. We isolate a narrower issue: training-data ancestry of the solver used for proposer feedback. We do not claim to introduce self-play, answer verification, or cross-fitting itself.

\noindent\textbf{Proxy rewards and self-confirmation.} Reward hacking and reward-process manipulation predate language agents \citep{amodei2016,everitt2019}. Proxy reward optimization can diverge from a ground-truth objective \citep{gao2022}; human-preference training can also favor agreement over truth \citep{sharma2023}. Pseudo-label confirmation bias provides a related account of learning from one's own mistakes \citep{arazo2020}. Our focus is the additional return path from a pseudo-label-trained solver into task generation. False agreement is a diagnostic for this path, not by itself a causal proof of exploitation. Automated judges can have systematic biases \citep{zheng2023judge}, motivating blinded evidence gathering and human validation.

\noindent\textbf{Data exclusion versus better evidence.} Cross-fitting uses held-out nuisance predictions in statistical estimation \citep{chernozhukov2018}. We borrow its exclusion principle, not its asymptotic guarantees: our adaptive curriculum lacks a demonstrated orthogonal score or independent sample structure. Retrieval and iterative search improve access to evidence \citep{lewis2020retrieval,guu2020retrieval,trivedi2023interleaving,li2025search}; they do not establish independence between a pseudo-label and a trained evaluator. Likewise, majority stability does not imply correctness when labelers share a model and evidence. These distinctions motivate evaluating verification and source exclusion as separate interventions. Appendix~\ref{app:literature} expands the literature map.

\vspace{10pt}
\section{Conclusion}
Co-cheating exposes a failure of self-evolution: agreement can improve because a proposer and solver reinforce the same incorrect labels. Our evidence-backed audit separates this internal progress from correctness. \texttt{CrossFit} addresses the feedback path by scoring each source with an auxiliary solver trained on the complementary fold, while the main solver still learns from all admitted tasks.

Across Qwen3.5-4B and Qwen3.5-9B, this change reduces final-round false agreement from 6.1\%/8.8\% to 3.0\%/3.7\% and improves seven-benchmark average Cover-EM over Dr.~Zero by 8.8/8.4 points. Fixed-bank replay and evaluator controls support source ancestry as the operative distinction; verification provides complementary reliability gains but only modest additional downstream improvement. These findings motivate tracking how an evaluator acquired its supervision when designing self-generated curricula. Shared pretraining errors, overlapping evidence, and auxiliary cost remain open limitations (Appendix~\ref{app:discussion-limitations}): extending exclusion to connected sources and measuring end-to-end efficiency are essential next tests of the principle. Reliable self-evolution therefore requires auditing both feedback correctness and the training history of its evaluator.

\FloatBarrier
\label{main-end}
\clearpage

\section*{AI use statement}
An AI coding and writing assistant assisted with literature discovery, draft organization, consistency checks, and LaTeX authoring. An LLM accessed through a commercial API (\texttt{gpt-6-astra/high}) served as the judge in the post-hoc audit described in Section~\ref{sec:audit} and Appendix~\ref{app:eval}. The authors verified the reported results, claims, citations, and implementation correspondence.

\section*{Reproducibility statement}
The appendix reports the training schedule, model configuration, evaluation manifest, scoring rules, audit procedure, and mechanism-specific controls used in our experiments.

\bibliography{references}
\bibliographystyle{iclr2027_conference}

\appendix
\section{Implementation details}
\label{app:impl}

\paragraph{Optimization.}
The proposer and solver are updated with a sequence-normalized policy-gradient objective. For usable trajectories $\mathcal E$ and scored assistant tokens $\mathcal T_e$,
\begin{equation}
 \mathcal L(\theta)=-\frac1{|\mathcal E|}\sum_{e\in\mathcal E}\widehat A_e
 \frac1{|\mathcal T_e|}\sum_{t\in\mathcal T_e}\log\pi_\theta(y_{e,t}\mid y_{e,<t},q_e),\qquad
 \widehat A_e=\frac{r_e-\mu_{g(e)}}{\sigma_{g(e)}+10^{-6}}.
 \label{eq:pg}
\end{equation}
Advantages are normalized within proposer task buckets or within the five solver responses to one question. Prompt and tool tokens are masked, as is the proposer's terminal answer; solver answer tokens remain trainable.

\paragraph{Training configuration.}
All treatments use the same backbone, proposer schedule, main-solver schedule, sampling configuration, and tool budget. The main solver trains on all admitted questions. In \texttt{CrossFit}, two auxiliary solvers train only on their assigned source folds and are used solely to produce proposer feedback on the complementary fold.

\begin{table}[t]
\raggedright
\small
\caption{Training configuration shared across treatments.}
\label{tab:config}
\begin{tabularx}{\linewidth}{@{}lX@{}}
\toprule
Item & Configuration\\
\midrule
Backbones & Qwen3.5-4B and Qwen3.5-9B\\
Schedule & Three rounds; 18 proposer and 25 main-solver updates per round\\
Solver data & 1,600 admitted questions per round; five responses per question\\
Proposer update & 64 prompts per update; one selected trajectory per prompt\\
Sampling & Temperature 0.8 and top-$p$ 0.95 for training trajectories\\
\texttt{MSV} & Three source-aware and three source-blind samples per proposal\\
\texttt{CrossFit} & Two source folds; 25 updates per round for each auxiliary solver (50 in total)\\
Tool budget & At most five assistant actions and 512 new tokens per action\\
\bottomrule
\end{tabularx}
\end{table}

\paragraph{Compute overhead.}
\texttt{MSV} adds six labeler generations per proposal, and \texttt{CrossFit} adds 50 auxiliary solver updates per round in the main experiments (25 per fold) without changing the main solver's training data or update count. Table~\ref{tab:cost} reports the resulting resource cost. Relative to Dr.~Zero, \texttt{MSV} raises the reserved budget by about 90\% at both scales (379 to 719 H200-hours on Qwen3.5-4B and 476 to 903 on Qwen3.5-9B) and multiplies judge requests almost fivefold, whereas \texttt{CrossFit} adds 72\% and 79\%. Halving its auxiliary budget to 25 updates per round lowers this to 36\% and 40\% with nearly the same replay false-agreement mass (0.005 versus 0.004 on Qwen3.5-4B and 0.002 versus 0.001 on Qwen3.5-9B; Figure~\ref{fig:ablation}). Combining both interventions costs 2.6 and 2.7 times the Dr.~Zero budget. Because reserved hours include waiting time, these ratios compare budgets rather than accelerator utilization.
\begin{table}[t]
\raggedright
\footnotesize
\caption{\textbf{Resource cost per training run.} H200-hours are reserved budgets: each run holds eight H200 GPUs for its end-to-end duration, including waiting time, so they exceed the accelerator time actually used; the GPU usage of the external audit service is unknown and not included. Token (millions) and judge-request (thousands) counts are per single-scale run, not summed over the two scales, and exclude training-replay tokens. Hours are the end-to-end wall-clock duration of the full pipeline. ``25 total'' and ``25 per fold'' denote the auxiliary-update budget per round.}
\label{tab:cost}
\setlength{\tabcolsep}{4pt}
\begin{tabular*}{\linewidth}{@{\extracolsep{\fill}}lrrrrrrr@{}}
\toprule
& \multicolumn{2}{c}{H200-hours} & \multicolumn{2}{c}{Tokens (M)} & Judge & \multicolumn{2}{c}{Hours} \\
\cmidrule(lr){2-3}\cmidrule(lr){4-5}\cmidrule(lr){7-8}
Treatment & 4B & 9B & Input & Output & req.\ (k) & 4B & 9B \\
\midrule
Dr.~Zero & 379 & 476 & 696 & 83 & 60.4 & 47 & 60 \\
\texttt{MSV} & 719 & 903 & 1,811 & 195 & 296.9 & 90 & 113 \\
\texttt{CrossFit} (25 total) & 515 & 665 & 889 & 107 & 87.1 & 64 & 83 \\
\texttt{CrossFit} (25 per fold) & 650 & 854 & 1,081 & 131 & 113.8 & 81 & 107 \\
\texttt{MSV}+\texttt{CrossFit} (25 per fold) & \textbf{990} & \textbf{1,281} & \textbf{2,196} & \textbf{243} & \textbf{350.3} & \textbf{124} & \textbf{160} \\
\bottomrule
\end{tabular*}
\end{table}

\section{Evaluation and audit details}
\label{app:eval}

\paragraph{Downstream evaluation.}
We extract the terminal answer, normalize case, punctuation, English articles, and whitespace, and score against the best matching reference alias. Cover-EM is one when a nonempty normalized reference answer appears in the normalized prediction. Every checkpoint is evaluated on the same 1,325-question manifest with one greedy search trajectory, an identical tool budget, and identical answer extraction. For benchmark $d$ with $n_d$ questions and scores $z_{di}$, we report
\begin{equation}
 \operatorname{Micro}=\frac{\sum_d\sum_{i=1}^{n_d}z_{di}}{1325},\qquad
 \operatorname{Macro}=\frac17\sum_{d=1}^{7}\frac1{n_d}\sum_{i=1}^{n_d}z_{di}.
 \label{eq:aggregate}
\end{equation}

\paragraph{Independent audit.}
Many generated questions require reconstructing multi-hop evidence across long or specialized source documents, making exhaustive human adjudication at every training step impractical. At every audited step, we therefore save the source document, adopted pseudo-label, and five solver responses used for proposer reward. \texttt{gpt-6-astra/high} constructs an evidence-backed reference from the source and judges the saved label and responses against that reference. Unsupported cases remain unresolved and are included in coverage accounting. The auditor is post-hoc: it never changes admission, model updates, or proposer reward. Table~\ref{tab:audit-abs} lists the round-3 mean of each audit statistic for every treatment.

\begin{table}[t]
\raggedright
\caption{Round-3 audit statistics for every treatment: means of the 43 round-3 step rates. $J/E$ is audit coverage; $T_P$, $T_S$, $A$, $F$, and $L$ are defined in Section~\ref{sec:audit}. Figure~\ref{fig:audit} plots the corresponding changes relative to Dr.~Zero, computed from unrounded means; they can therefore differ by 0.1 percentage point from differences of the rounded entries here.}
\small
\begin{tabular*}{\linewidth}{@{\extracolsep{\fill}}lcccccc@{}}
\toprule
\multirow{2}{*}{\textbf{}} & $J/E$ & $T_P$ & $T_S$ & $A$ & $F$ & $L$ \\ \cmidrule(l){2-7}
& \multicolumn{6}{c}{Qwen3.5-4B} \\ \midrule
\textbf{Dr.~Zero} & 0.859 & 0.747 & 0.669 & 0.710 & 0.061 & 0.020 \\
\textbf{\texttt{MSV}} & 0.860 & 0.747 & 0.708 & 0.745 & 0.057 & 0.020 \\
\textbf{\texttt{CrossFit}} & 0.860 & 0.819 & 0.686 & 0.679 & 0.030 & 0.038 \\
\textbf{\texttt{MSV} + \texttt{CrossFit}} & 0.858 & 0.843 & 0.727 & 0.708 & 0.020 & 0.038 \\ \cmidrule(l){2-7}
\multicolumn{1}{c}{\textbf{}} & \multicolumn{6}{c}{Qwen3.5-9B} \\ \midrule
\textbf{Dr.~Zero} & 0.859 & 0.737 & 0.689 & 0.752 & 0.088 & 0.025 \\
\textbf{\texttt{MSV}} & 0.853 & 0.770 & 0.727 & 0.788 & 0.072 & 0.011 \\
\textbf{\texttt{CrossFit}} & 0.857 & 0.851 & 0.712 & 0.709 & 0.037 & 0.041 \\
\textbf{\texttt{MSV} + \texttt{CrossFit}} & 0.861 & 0.878 & 0.754 & 0.732 & 0.017 & 0.039 \\ \bottomrule
\end{tabular*}
\label{tab:audit-abs}
\end{table}

\paragraph{Round-wise results.}
Table~\ref{tab:downstream-rounds} lists Cover-EM for every round-end main solver in the layout of Table~\ref{tab:downstream}, and Table~\ref{tab:micro} gives the corresponding micro averages, which weight all 1,325 questions equally. Base obtains micro averages of 0.382 on Qwen3.5-4B and 0.404 on Qwen3.5-9B. Micro and macro averages order the treatments identically in every round.

\begin{table*}[t]
\raggedright
\caption{Cover-EM of every round-end main solver; we mark the best performance within each scale in bold. Each cross-fitted treatment shares round 1 with its coupled counterpart.}
\resizebox{\linewidth}{!}{
\begin{tabular}{@{}lcccccccc@{}}
\toprule
\multirow{2}{*}{\textbf{}}                        & \textbf{NQ}    & \textbf{TriviaQA} & \textbf{PopQA} & \textbf{HotpotQA} & \textbf{2WikiMQA} & \textbf{MuSiQue} & \textbf{Bamboogle} & \textbf{Average} \\ \cmidrule(l){2-9} 
                                                  & \multicolumn{8}{c}{Qwen3.5-4B}                                                                                                                         \\ \midrule
\textbf{Base}                                     & 0.380          & 0.655             & 0.315          & 0.350             & 0.465             & 0.105            & 0.416              & 0.384            \\
\textbf{Dr.~Zero Round 1}                          & 0.380          & 0.660             & 0.325          & 0.350             & 0.475             & 0.125            & 0.424              & 0.391            \\
\textbf{Dr.~Zero Round 2}                          & 0.395          & 0.665             & 0.320          & 0.360             & 0.470             & 0.125            & 0.440              & 0.396            \\
\textbf{Dr.~Zero Round 3}                          & 0.390          & 0.665             & 0.325          & 0.365             & 0.485             & 0.120            & 0.448              & 0.400            \\
\textbf{\texttt{MSV} Round 1}                     & 0.385          & 0.675             & 0.330          & 0.355             & 0.480             & 0.110            & 0.432              & 0.395            \\
\textbf{\texttt{MSV} Round 2}                     & 0.390          & 0.670             & 0.335          & 0.370             & 0.490             & 0.120            & 0.432              & 0.401            \\
\textbf{\texttt{MSV} Round 3}                     & 0.400          & 0.670             & 0.340          & 0.375             & 0.480             & 0.135            & 0.448              & 0.407            \\
\textbf{\texttt{CrossFit} Round 1}                & 0.380          & 0.660             & 0.325          & 0.350             & 0.475             & 0.125            & 0.424              & 0.391            \\
\textbf{\texttt{CrossFit} Round 2}                & 0.415          & 0.685             & 0.370          & 0.420             & 0.535             & 0.165            & 0.480              & 0.439            \\
\textbf{\texttt{CrossFit} Round 3}                & 0.455          & \textbf{0.730}    & \textbf{0.415} & 0.460             & \textbf{0.575}    & 0.245            & \textbf{0.536}     & 0.488            \\
\textbf{\texttt{MSV} + \texttt{CrossFit} Round 1} & 0.385          & 0.675             & 0.330          & 0.355             & 0.480             & 0.110            & 0.432              & 0.395            \\
\textbf{\texttt{MSV} + \texttt{CrossFit} Round 2} & 0.440          & 0.680             & 0.370          & 0.420             & 0.515             & 0.205            & 0.496              & 0.447            \\
\textbf{\texttt{MSV} + \texttt{CrossFit} Round 3} & \textbf{0.470} & \textbf{0.730}    & 0.410          & \textbf{0.475}    & \textbf{0.575}    & \textbf{0.250}   & 0.528              & \textbf{0.491}   \\ \cmidrule(l){2-9} 
\multicolumn{1}{c}{\textbf{}}                     & \multicolumn{8}{c}{Qwen3.5-9B}                                                                                                                         \\ \midrule
\textbf{Base}                                     & 0.505          & 0.710             & 0.375          & 0.355             & 0.295             & 0.125            & 0.496              & 0.409            \\
\textbf{Dr.~Zero Round 1}                          & 0.500          & 0.725             & 0.385          & 0.365             & 0.310             & 0.140            & 0.496              & 0.417            \\
\textbf{Dr.~Zero Round 2}                          & 0.520          & 0.725             & 0.385          & 0.370             & 0.310             & 0.135            & 0.520              & 0.424            \\
\textbf{Dr.~Zero Round 3}                          & 0.515          & 0.730             & 0.400          & 0.370             & 0.320             & 0.150            & 0.512              & 0.428            \\
\textbf{\texttt{MSV} Round 1}                     & 0.510          & 0.725             & 0.390          & 0.365             & 0.305             & 0.150            & 0.504              & 0.421            \\
\textbf{\texttt{MSV} Round 2}                     & 0.525          & 0.710             & 0.395          & 0.375             & 0.320             & 0.150            & 0.528              & 0.429            \\
\textbf{\texttt{MSV} Round 3}                     & 0.525          & 0.720             & 0.400          & 0.390             & 0.325             & 0.155            & 0.536              & 0.436            \\
\textbf{\texttt{CrossFit} Round 1}                & 0.500          & 0.725             & 0.385          & 0.365             & 0.310             & 0.140            & 0.496              & 0.417            \\
\textbf{\texttt{CrossFit} Round 2}                & 0.550          & 0.750             & 0.430          & 0.430             & 0.355             & 0.200            & 0.552              & 0.467            \\
\textbf{\texttt{CrossFit} Round 3}                & \textbf{0.580} & 0.765             & 0.455          & \textbf{0.490}    & \textbf{0.425}    & 0.255            & \textbf{0.616}     & 0.512            \\
\textbf{\texttt{MSV} + \texttt{CrossFit} Round 1} & 0.510          & 0.725             & 0.390          & 0.365             & 0.305             & 0.150            & 0.504              & 0.421            \\
\textbf{\texttt{MSV} + \texttt{CrossFit} Round 2} & 0.550          & 0.745             & 0.435          & 0.425             & 0.365             & 0.220            & 0.576              & 0.474            \\
\textbf{\texttt{MSV} + \texttt{CrossFit} Round 3} & 0.570          & \textbf{0.780}    & \textbf{0.470} & 0.485             & 0.415             & \textbf{0.280}   & 0.608              & \textbf{0.515}   \\ \bottomrule
\end{tabular}
}
\label{tab:downstream-rounds}
\end{table*}

\begin{table}[t]
\raggedright
\caption{Micro-averaged Cover-EM of the round-end main solvers; we mark the best performance in bold.}
\small
\begin{tabular*}{\linewidth}{@{\extracolsep{\fill}}lcccccc@{}}
\toprule
\multirow{2}{*}{\textbf{}}                & \multicolumn{3}{c}{Qwen3.5-4B}                         & \multicolumn{3}{c}{Qwen3.5-9B}                         \\ \cmidrule(l){2-4} \cmidrule(l){5-7} 
                                          & \textbf{Round 1} & \textbf{Round 2} & \textbf{Round 3} & \textbf{Round 1} & \textbf{Round 2} & \textbf{Round 3} \\ \midrule
\textbf{Dr.~Zero}                          & 0.389            & 0.394            & 0.397            & 0.413            & 0.418            & 0.423            \\
\textbf{\texttt{MSV}}                     & \textbf{0.393}   & 0.399            & 0.405            & \textbf{0.417}   & 0.423            & 0.430            \\
\textbf{\texttt{CrossFit}}                & 0.389            & 0.436            & 0.485            & 0.413            & 0.462            & 0.506            \\
\textbf{\texttt{MSV} + \texttt{CrossFit}} & \textbf{0.393}   & \textbf{0.444}   & \textbf{0.489}   & \textbf{0.417}   & \textbf{0.468}   & \textbf{0.510}   \\ \bottomrule
\end{tabular*}
\label{tab:micro}
\end{table}

\section{Broader literature map}
\label{app:literature}
This map separates complementary research questions rather than treating every cited system as a direct experimental baseline. Only methods with accessible implementations and matched protocols can support comparative performance claims.

\paragraph{Retrieval representations and evidence use.}
Dense retrieval, fusion-in-decoder, retrieval-enhanced language modeling, and few-shot retrieval pretraining study how evidence is retrieved and represented \citep{karpukhin2020dense,izacard2021leveraging,borgeaud2022improving,izacard2023atlas}. Query rewriting and active retrieval change when and how evidence is requested \citep{ma2023query,jiang2023active}. Black-box retrieval augmentation, hierarchical retrieval, and chain-of-note processing provide other evidence interfaces \citep{shi2024replug,sarthi2024raptor,yu2023chain}. Robustness to irrelevant context, corrective retrieval, and retrieval-aware tuning address evidence quality or utilization \citep{yoran2023making,yan2024corrective,lin2024ra}. These directions motivate holding the retrieval backend fixed: a change in evidence access must not be mistaken for an effect of source-excluded feedback.

\paragraph{Search-agent optimization.}
Beyond Search-R1 and R1-Searcher, evolving search, efficient search training, simulated search, deep-research agents, and trained web agents illustrate the expanding space of agent learning \citep{zhang2025evolvesearch,jiang2025s3,sun2025zerosearch,zheng2025deepresearcher,zhang2025web}. Step-level search training and systems for difficult web exploration further motivate recording trajectory budgets and tool usage \citep{wang2025stepsearch,li2025websailor,liu2025webexplorer}. Toolformer studies learning tool use, while more recent open research and search-agent frameworks expand the surrounding system design space \citep{schick2023toolformer,li2026openresearcher,chu2026redsearcher,liang2026search}. They are contextual references, not claims of matched evaluation in this manuscript.

\paragraph{Self-generated learning and reinforcement learning.}
Self-play, multi-agent games, self-training, and iterative reasoning optimization offer different sources of automatically generated supervision \citep{ye2024scalable,liu2025spiral,zhang2024rest,pang2024iterative}. Outside language modeling, likelihood-free inference also learns from generated supervision, fitting reconstruction maps or posterior densities to simulator samples whose parameters are known by construction \citep{zhang2025likelihoodfree,zhang2025reconstruction}; the generated pairs studied here differ in that their labels are themselves model predictions, so label error can propagate into the evaluator. Preference learning and large-scale reasoning RL establish additional optimization choices \citep{ouyang2022training,rafailov2023direct,guo2025deepseek,yu2025dapo}. Our inspected sequence-level policy-gradient implementation is specified directly in our implementation audit; citing these methods does not imply that their algorithms, training data, or reported capabilities are reproduced. The distinctive variable studied here is the ancestry of the policy supplying proposal feedback, not a new generic policy-gradient estimator.

\section{Discussion and limitations}
\label{app:discussion-limitations}
Cross-fitting changes who supplies feedback, not what makes an answer true. Its source exclusion is valuable only if checkpoint ancestry and data routing enforce it. Shared pretraining, overlapping web evidence, semantically related sources, and an adaptive proposer can still induce correlated mistakes. A lower false-agreement mass on accepted questions can also result from rejecting difficult tasks rather than improving learning. Coverage, task difficulty, fixed-probe performance, and downstream capability must therefore accompany that number. Our results establish empirical mitigation across two model scales, but do not yet establish lower end-to-end cost or robustness to connected sources.

\section{Complete per-step audit trajectories}
\label{app:complete-audit-trajectories}
\begin{figure}[H]
\raggedright
\includegraphics[width=\linewidth]{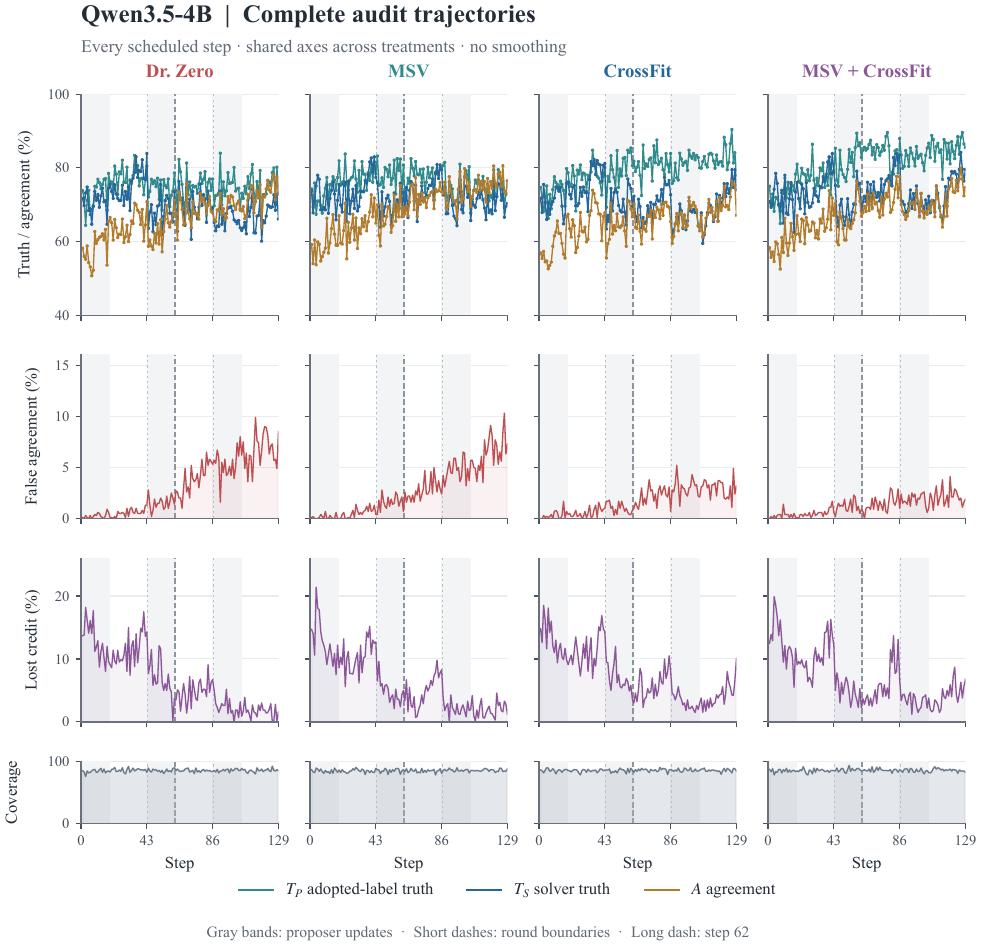}
\caption{\textbf{Complete Qwen3.5-4B audit trajectories.} Columns identify treatments. Rows show adopted-label truth $T_P$, solver truth $T_S$, and agreement $A$; false-agreement mass $F$; lost-credit mass $L$; and coverage $J/E$. Every metric is expressed in percent. Shared limits support comparison across treatments and model scales. Gray bands mark proposer phases, short dashes mark round boundaries, and the long dash marks step 62.}
\label{fig:audit-full-4b}
\end{figure}

\begin{figure}[H]
\raggedright
\includegraphics[width=\linewidth]{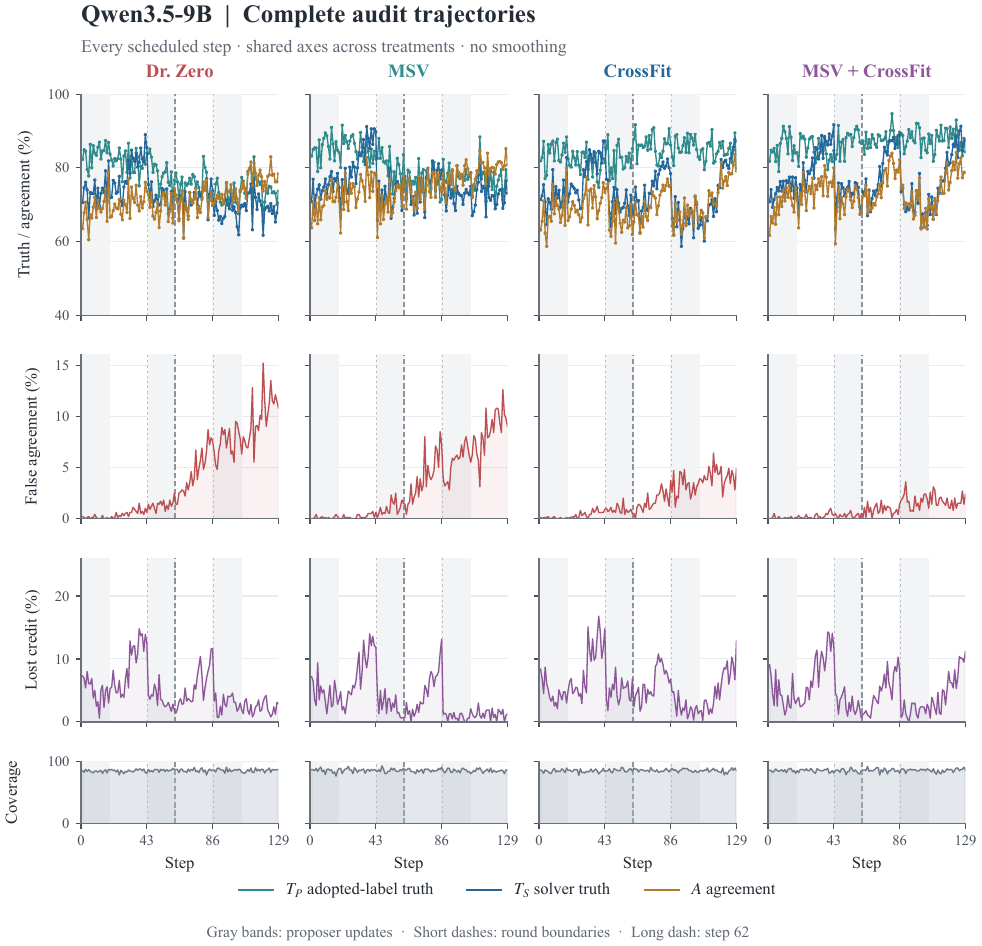}
\caption{\textbf{Complete Qwen3.5-9B audit trajectories.} Layout, metric colors, units, and axis limits match Figure~\ref{fig:audit-full-4b}. Every scheduled step is retained. The separate coverage strip prevents audit coverage from obscuring the correctness curves.}
\label{fig:audit-full-9b}
\end{figure}

Figures~\ref{fig:audit-full-4b} and~\ref{fig:audit-full-9b} retain all 129 scheduled steps, all four treatments, and all six metrics from the data underlying Figure~\ref{fig:dynamics}. Values are plotted directly from the existing per-step table, in percent, without smoothing or interpolation. Separate axes for false agreement and lost credit keep their distinct magnitudes visible; coverage is the audited fraction $J/E$ and is displayed separately from correctness. The round summaries in Figure~\ref{fig:dynamics} are arithmetic means of the 43 displayed step rates in each round, not additional runs or uncertainty estimates.

\clearpage

\clearpage

\end{document}